\documentclass{article}
\usepackage{spconf,amsmath,epsfig}
\usepackage{algorithm}
\usepackage{algorithmic}
\usepackage{amsfonts}
\usepackage{subfigure}
\usepackage{color}
\begin{document}\sloppy
\onecolumn
\def\x{{\mathbf x}}
\def\L{{\cal L}}

\title{\Large S\lowercase{ync}GAN: S\lowercase{ynchronize} \lowercase{the} l\lowercase{atent} s\lowercase{pace} \lowercase{of} c\lowercase{ross-modal} \\g\lowercase{enerative} a\lowercase{dversarial} N\lowercase{etworks}}

\name{Wen-Cheng Chen, Chien-Wen Chen, and Min-Chun Hu}
\address{Dept. of Computer Science and Information Engineering, \\
	National Cheng Kung University, Tainan 701, Taiwan. \\
    \{jerrywiston, ai281918, trimy\}@mislab.csie.ncku.edu.tw}

\maketitle
\begin{abstract}
\textit{
Generative adversarial network (GAN) has achieved impressive success on cross-domain generation, but it faces difficulty in cross-modal generation due to the lack of a common distribution between heterogeneous data. Most existing methods of conditional based cross-modal GANs adopt the strategy of one-directional transfer and have achieved preliminary success on text-to-image transfer. Instead of learning the transfer between different modalities, we aim to learn a synchronous latent space representing the cross-modal common concept. A novel network component named synchronizer is proposed in this work to judge whether the paired data is synchronous/corresponding or not, which can constrain the latent space of generators in the GANs. Our GAN model, named as SyncGAN, can successfully generate synchronous data (e.g., a pair of image and sound) from identical random noise. For transforming data from one modality to another, we recover the latent code by inverting the mappings of a generator and use it to generate data of different modality. In addition, the proposed model can achieve semi-supervised learning, which makes our model more flexible for practical applications.
}
\end{abstract}
\section{Introduction}
\label{sec:intro}
Every concept in the world can be presented in many different digital modalities such as an image, a video, a clip of sound, or a few text. 
In recent years, many researchers in the multimedia field have been devoted to \textbf{cross-domain} generation, which aims to generate data of the same modality but with different kinds of representations/styles. For example, transferring a photo to an art painting. Moreover, some researchers try to generate \textbf{cross-modal} data. For example, given a video of violin playing, generating an audio clip of violin sound. In general, a pair of \textbf{cross-domain} data show different representations/styles of a concept, but the paired data still have a common shape structure. On the other hand, a pair of \textbf{cross-modal} data usually have heterogeneous features with quite different distributions, and therefore it is much more challenging to model the relationship between cross-modal data.

Deep learning has made tremendous progress on style transfer and cross-domain generation. For example, fully convolutional network (FCN)~\cite{long2015fully} was first proposed to achieve image to image transformation by using inverse convolutional layer. It can successfully process an input image into an output segmentation result. Isola \textit{et al}. introduced Pix2Pix~\cite{pix2pix2017}, which combines conditional GAN and L1/L2 distance loss of paired data to generate images with more clear visual content. Moreover, DiscoGAN~\cite{kim2017learning}, CycleGAN~\cite{zhu2017unpaired} and DualGAN~\cite{yi2017dualgan} achieved unsupervised cross-domain transformation (i.e. the training process can be done without paired data) by using cycle-consistency structure to find the common distribution. Coupled GAN~\cite{liu2016coupled} performed cross-domain paired data generation by coupling two GANs with shared weightings. As for cross-modal generation, Reed \textit{et al}.~\cite{reed2016generative} successfully generated images from text by using conditional GAN and a matching-aware discriminator. Chen \textit{et al}.~\cite{chen2017deep} used the same method to transfer between sound and images. All of the above cross-modal GANs are conditional based; therefore, they can only perform one-directional transformation. That is, these GANs cannot generate a pair of synchronous cross-modal data simultaneously. 

In this work, we propose a new GAN model named SyncGAN, which can learn a synchronous latent space representing the cross-domain or cross-modal data. 
The main contributions of this work are summarized as follows:
(1) A synchronizer is introduced to estimate the synchronous probability and constrain the latent space of generators using paired data without any class label. (2) Our SyncGAN model can successfully generate synchronous data from identical random noise. Moreover, the latent code of data in one modality can be recovered by inverting the mappings of the generator, and the corresponding data in another modality can be generated based on the latent code. (3) The proposed SyncGAN can also achieve semi-supervised learning by using a small amount of training data with information of synchronous/asynchronous, which makes our model more flexible for practical applications. To the best of our knowledge, SyncGAN is the first generative model which can perform synchronous data generation and bidirectional modality transformation.

\section{Model description}
\label{sec:Model}
\begin{figure}[!t]
	\centering
	\includegraphics[width=4in]{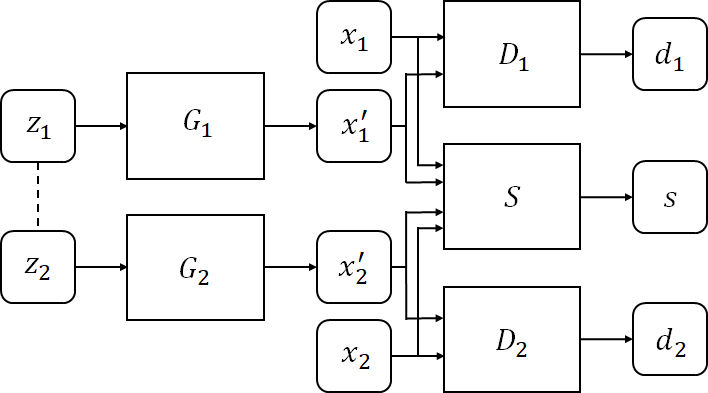}\\
	\caption{The structure of the proposed SyncGAN model.}\label{SyncGAN}
\end{figure}

\subsection{Generative Adversarial Nets}
Generative Adversarial Nets~\cite{goodfellow2014generative} are widely applied in generative models and have outstanding performance. The main concept of the Generative Adversarial Nets is the adversarial relationship between two models: a generative model $G$ and a discriminative model $D$. The objective of $G$ is to capture the distribution of training data, and $D$ estimates the probability that a sample came from the training data rather than the generator $G$. With training data $x$ and sampled noise $z$, $D$ is used to maximize $log(D(x)) + log(1-D(G(z)))$ and $G$ is for minimizing $log(1-D(G(z)))$. $D$ and $G$ play the following minimax game with the value function $V(G,D)$:
\begin{eqnarray}
\label{eq:gan}
\displaystyle \min_{G} \max_{D} V(D,G)=\mathbb{E}_{x \sim p_{data}(x)}[\log D(x)]\nonumber
+\mathbb{E}_{z \sim p_{z}(z)}[\log (1-D(G(z))))]
\end{eqnarray}
\subsection{Synchronous Generative Adversarial Nets}
Fig.~\ref{SyncGAN} illustrates the architecture of the proposed SyncGAN model, which is constructed based on two general GANs ($G_1, D_1$) and ($G_2, D_2$). Let $m$ indicates the data modality, each generator $G_m$ is used to generate data of  modality $m$ (i.e. $x_m'$) and each discriminator $D_m$ is used to estimate the probability $d_m$ that a sample is real data of modality $m$. Moreover, two generators are connected to an additional network named synchronizer ($S$), which is used to synchronize the latent space of two generators. The details of each block are described as follows.\newline\newline
\textbf{Discriminators:} Our discriminators work in a similar way as the general GANs. The objective of a discriminator is to estimate the probability that a sample came from the training data rather than the generator $G_m$ by maximizing the loss function defined in Eq.~\ref{eq:sync_d}. In this loss function, $x_m$ is the real data from the distribution $p(x_m)$, and $z_m$ is the latent vector of the modality $m$. Both $z_1$ and $z_2$ are sampled from the same distribution $p(z)$.

\begin{eqnarray}
\label{eq:sync_d}
\mathcal{L}_{D_m}=\mathbb{E}_{x_m\sim p(x_m)}[\log D_m(x_m))]\nonumber+\mathbb{E}_{z_m\sim p(z)}[log(1-D_m(G_m(z_m)))]
\end{eqnarray}
\textbf{Synchronizer:} The synchronizer takes the data from different modalities as input. The objective of the synchronizer is to estimate the probability that two input data are synchronous (i.e. representing the same concept). We use synchronous and asynchronous real data to train the synchronizer by maximizing the loss function defined in Eq.~\ref{eq:sync_s}. In this loss function, $i$ and $j$ denote the IDs of pairwise data in modality 1 and modality 2, respectively. 
Two synchronous data will have the same ID, while two different IDs imply that the two data are asynchronous.
\begin{figure}[!t]
	\centering 
	\subfigure[]{
		\includegraphics[width=3.6in]{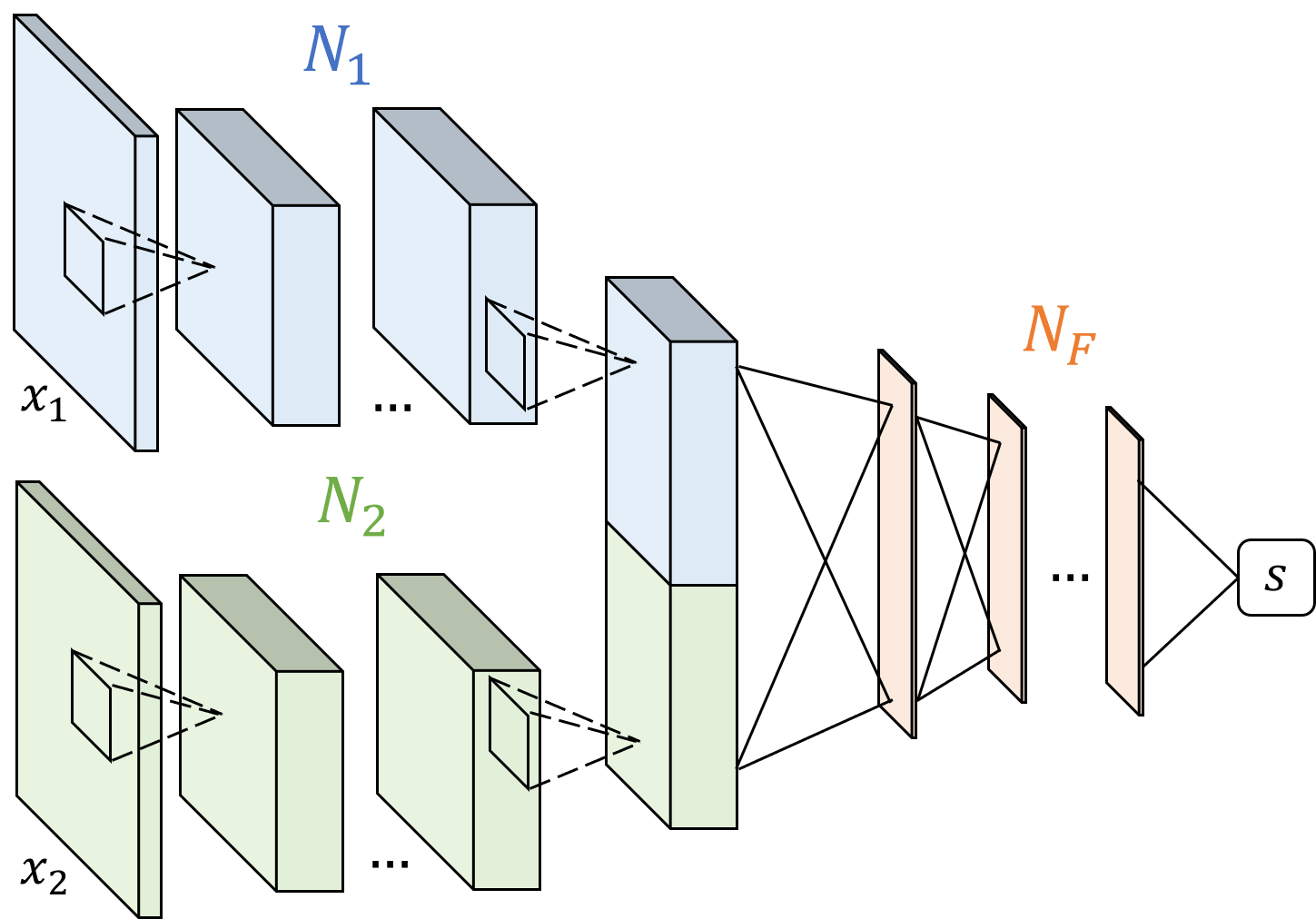}} 
	\subfigure[]{
		\includegraphics[width=3.3in]{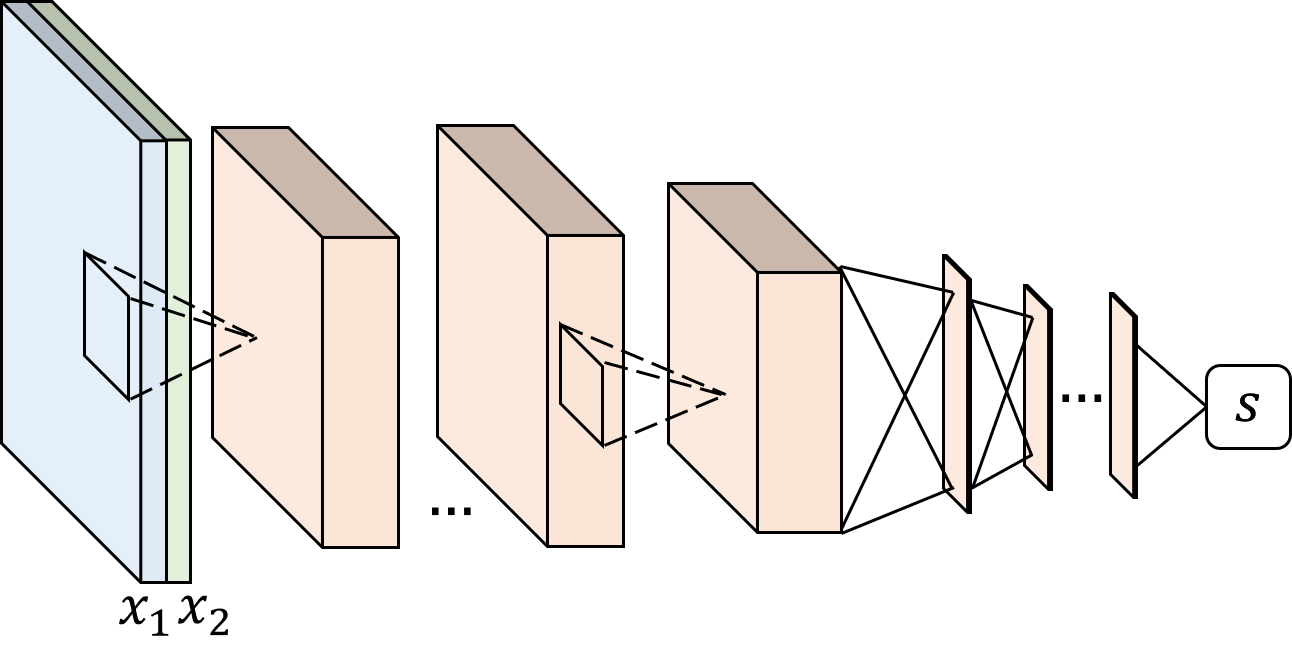}}
	\caption{Structure of the synchronizer. (a) is for cross-modal tasks and (b) is for style-transfer tasks.}
	\label{Synchronizer}
\end{figure}
\begin{eqnarray}
\label{eq:sync_s}
\mathcal{L}_S=\mathbb{E}_{x_1\sim p(x_1),x_2 \sim p(x_2)}[log(S(x_1^i,x_2^j))|i=j] \nonumber+ \mathbb{E}_{x_1\sim p(x_1),x_2 \sim p(x_2)}[log(1-S(x_1^i,x_2^j))|i \neq j]
\end{eqnarray}
Fig.~\ref{Synchronizer} shows the structure of the synchronizer $S$. Note that we design different synchronizer structures for different tasks. For the cross-modal data generation tasks, each data $x_m$ has its own network $N_m$ for feature extraction, and the extracted feature maps will be concatenated as the input of a network $N_F$ composed of fully connected layers (Fig.~\ref{Synchronizer}a). For the style-transfer tasks, data from different domains will be concatenated directly as the input of the synchronizer network (Fig.~\ref{Synchronizer}b).
\newline\newline
\textbf{Generators:} We use the noise $z_m$ sampled from normal distribution as the input of the generator $G_m$. Two generators are trained to capture the distribution of their target modalities. 
Like most existing GAN methods, the goal of the generators is to fool the discriminators by maximize the loss function $L_{G_m}^{Dis}$ defined in Eq.~\ref{eq:sync_gd}. Moreover, to constrain the latent space of generators, we should also consider the synchronous loss $L_{G}^{Sync}$ defined in Eq.~\ref{eq:sync_gs}. To be more precise, the network tends to generate synchronous data when the same noise is used for $z_1$ and $z_2$. In addition, asynchronous data would be generated when different noises are used for $z_1$ and $z_2$.
\begin{eqnarray}
\label{eq:sync_gd}
&&\mathcal{L}_{G_m}^{Dis}=\mathbb{E}_{z_m\sim p(z)}[log(D_m(G_m(z_m)))]\\
\label{eq:sync_gs}
&&\mathcal{L}_{G}^{Sync}=\mathbb{E}_{z_1,z_2\sim p(z)}[log(S(G_1(z_1),G_2(z_2)))|z_1=z_2]
+\mathbb{E}_{z_1,z_2\sim p(z)}[1-log(S(G_1(z_1),G_2(z_2)))|z_1 \neq z_2]
\end{eqnarray}
\begin{algorithm}
\caption{Training the SyncGAN model}
\label{train_SyncGAN}
\begin{algorithmic} 
\STATE $\theta_{S},\theta_{D_1},\theta_{D_2},\theta_{G_1},\theta_{G_2} \leftarrow \text{initialize network parameters}$
\REPEAT
\STATE $\text{// Data distribution loss}$
\STATE $z_1,z_2 \leftarrow p(z)$
\STATE $x_1,x_2 \leftarrow \text{random mini-batch from unpaired data}$
\STATE $\mathcal{L}_{D_1}\leftarrow \mathbb{E}_{x_1\sim p(x_1),z_1\sim p(z)}[log(D_1(x_1))+log(1-D_1(G_1(z_1)))]$
\STATE $\mathcal{L}_{D_2}\leftarrow \mathbb{E}_{x_2\sim p(x_2),z_2\sim p(z)}[log(D_2(x_2))+log(1-D_2(G_2(z_2)))]$
\STATE $\mathcal{L}_{G_1}^{Dis} \leftarrow \mathbb{E}_{z_1\sim p(z)}[log(D_1(G_1(z_1)))]$
\STATE $\mathcal{L}_{G_2}^{Dis} \leftarrow \mathbb{E}_{z_2\sim p(z)}[log(D_2(G_2(z_2)))]$
\STATE $ $
\STATE $\text{// Synchronous loss}$
\STATE $z_1',z_2' \leftarrow p(z)$
\STATE $x_1^i,x_2^j \leftarrow \text{random mini-batch from paired data}$
\STATE $\mathcal{L}_S \leftarrow \mathbb{E}_{x_1\sim p(x_1),x_2 \sim p(x_2)}[log(S(x_1^i,x_2^j))|\tiny{i=j}]+\mathbb{E}_{x_1\sim p(x_1),x_2 \sim p(x_2)}[log(1-S(x_1^i,x_2^j)|\tiny{i \neq j}]$
\STATE $\mathcal{L}_{G}^{Sync}\leftarrow\mathbb{E}_{z_1',z_2'\sim p(z)}[log(S(G_1(z_1'),G_2(z_2')))|\tiny{z_1'=z_2'}]+\mathbb{E}_{z_1',z_2'\sim p(z)}[log(1-S(G_1(z_1'),G_2(z_2')))|\tiny{z_1' \neq z_2'}]$
\STATE $ $
\STATE $\text{// Update network parameters}$
\STATE $\theta_{S} \stackrel{+}{\leftarrow} -\nabla_{\theta_{S}} \mathcal{L}_{S}$
\STATE $\theta_{D_1} \stackrel{+}{\leftarrow} -\nabla_{\theta_{D_1}} \mathcal{L}_{D_1}$
\STATE $\theta_{D_2} \stackrel{+}{\leftarrow} -\nabla_{\theta_{D_2}} \mathcal{L}_{D_2}$
\STATE $\theta_{G_1} \stackrel{+}{\leftarrow} -\nabla_{\theta_{G_1}}( \mathcal{L}_{G_1}^{Dis}+\mathcal{L}_{G}^{Sync})$
\STATE $\theta_{G_2} \stackrel{+}{\leftarrow} -\nabla_{\theta_{G_2}}( \mathcal{L}_{G_2}^{Dis}+\mathcal{L}_{G}^{Sync})$
\UNTIL{convergence}
\end{algorithmic}
\end{algorithm}
Note that in the training process, if we only consider the data sampled with $z_1=z_2$, the mode collapse issue will be serious. It is important to also sample data with $z_1\neq z_2$. In this work, the ratio of identical and distinct pairs used for training the synchronizer is 0.5.

\subsection{Semi-supervised Training}
Based on the loss functions defined in Eq.~\ref{eq:sync_gd} and Eq.~\ref{eq:sync_gs}, we can train the synchronizer and discriminator independently. In other words, the SyncGAN can achieve semi-supervised learning by using a small amount of training data with information of synchronous/asynchronous to learn the synchronous concept and a large amount of training data without information of synchronous/asynchronous to learn the data distribution. During an iteration of the training stage, $z_1$ and $z_2$ are sampled from a normal distribution, and unpaired data $x_1$ and $x_2$ are used to compute the loss related to data distribution, i.e., $\mathcal{L}_{G_m}^{Dis}$ and $\mathcal{L}_{D_m}$. We then construct another training batch by concatenating synchronous/asynchronous latent vector and data with information of synchronous/asynchronous to compute synchronous correlated loss, i.e., $\mathcal{L}_{G}^{Sync}$ and $\mathcal{L}_{S}$. Finally we update the network parameters of each part. Please refer to (Alg.~\ref{train_SyncGAN}) for the overall training procedure.

\section{Experiments}
\label{sec:Experiments}
To validate the proposed SyncGAN model, we conducted experiments on several datasets: MNIST \cite{lecun1998gradient}, Fashion-MNIST \cite{xiao2017fashion}, UT Zappos50K \cite{finegrained}\cite{semjitter} and an instrument dataset collected by ourselves. We perform cross-modal and cross-domain data generation on these datasets and the details are described in Section 3.1 and Section 3.2, respectively. We also used our model to transfer data between different modalities/domains and the results are shown in Section 3.3. 
Section 3.4 evaluates the synchronous rate of our model.

\subsection{Cross-modal Generation}
\begin{figure}[!t]
	\centering
	\includegraphics[width=4in]{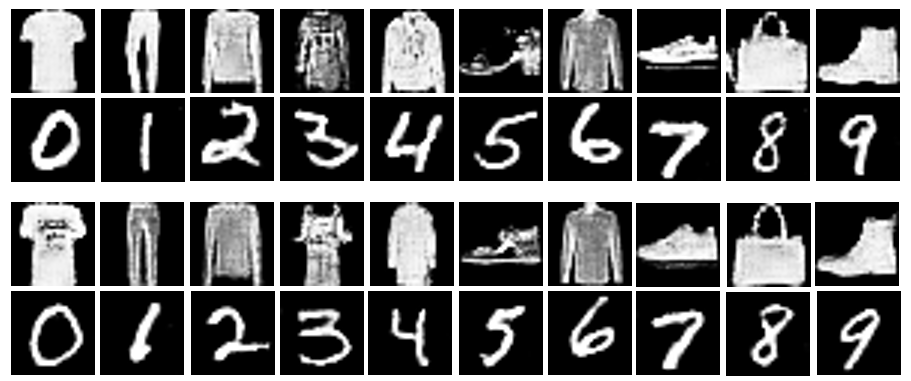}\\
	\caption{Synchronous generation of MNIST and Fashion-MNIST.}
    \label{sync_mnist_gen}
\end{figure}

\begin{figure}[!t]
	\centering
	\includegraphics[width=4in]{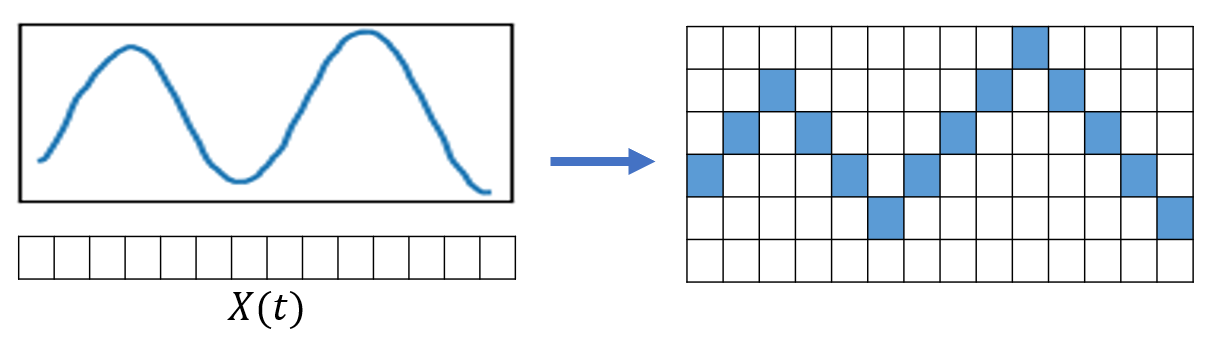}\\
	\caption{Illustration of the audio processing procedure.}
    \label{wav_processing}
\end{figure}

\begin{figure}[!t]
	\centering
	\includegraphics[width=4in]{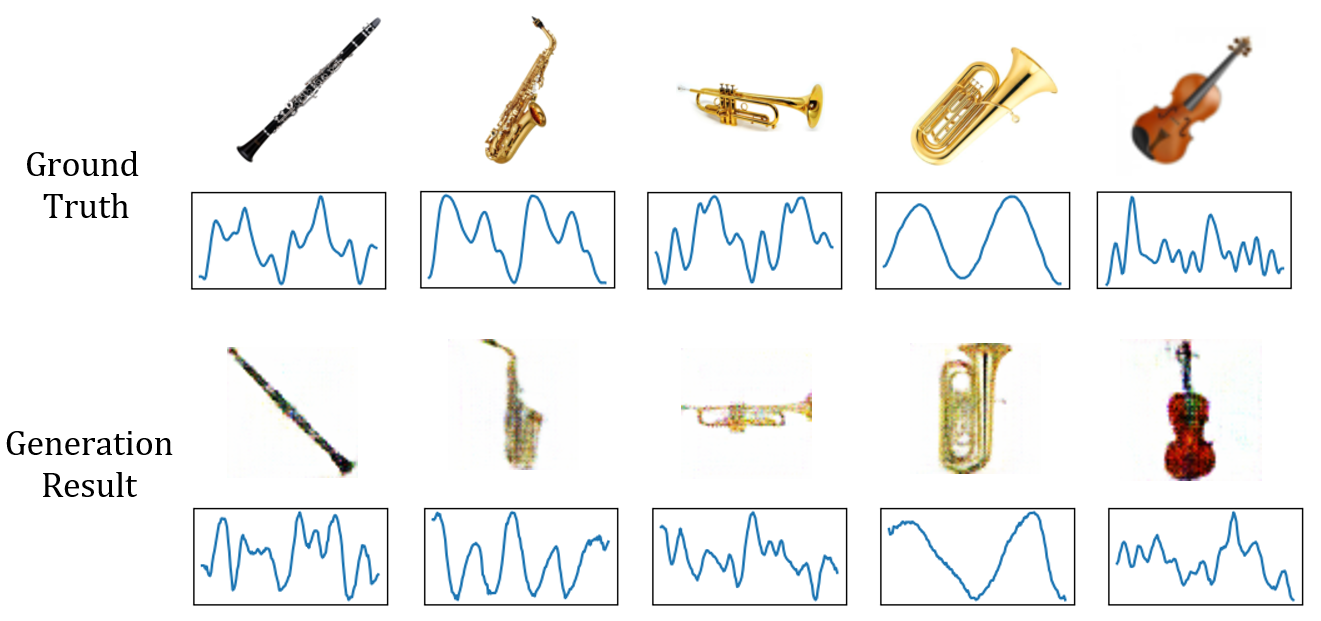}\\
	\caption{Synchronous generation of images and sound for instruments.}
    \label{img_wav}
\end{figure}

\textbf{MNIST and Fashion-MNIST:} The MNIST and Fashion-MNIST datasets contain ten image classes with 28x28 grayscale images of handwritten digits and clothes, respectively. Since the digits and the clothes do not have similar structure, we can consider the two datasets as two different modalities even though they are both image datasets. We used 30000 pairs of synchronous data from these two datasets for experiments. According to Table~\ref{tb:digit_fashion}, each paired data is obtained by sampling an image from class $Ci$ of the MNIST dataset and another image from also the class $Ci$ of the Fashion-MNIST dataset. Fig.~\ref{sync_mnist_gen}(a) shows that our SyncGAN model successfully generated image pairs of corresponding digit and cloth. Note that our model only needs to know the correspondence between two data rather than exact class labels.\newline\newline
\textbf{Image and audio of instruments:} We also conducted experiments on a cross-modal dataset containing images and audio clips of 5 kinds of instruments, including violin, trumpet, tuba, clarinet and sax. Each kind of instruments has 250 images and corresponding audio clips. Every image was cropped and normalized to 64x64 pixels. For audio data, we randomly clipped 512 samples from the wave file and down-sampled these samples to 128 values, denoted as $X(t), t=0,...127$. We further transformed the 1D sequence $X(t)$ to a 2D form with dimension of 64x128 as illustrated in Fig.~\ref{wav_processing}. Fig.~\ref{img_wav} shows the results of paired cross-modal data generated by our SyncGAN model. The generated audio waves are well synchronous with the generated images and we will show the synchronous rate in Section 3.4.

\begin{table}[t]
\begin{center}
\begin{tabular}{|c|c|c|}
  \hline
  Class Index & MNIST & Fashion-MNIST
  \\
  \hline
  C0 & 0 & T-shirt/top \\
  C1 & 1 & Trouser \\
  C2 & 2 & Pullover \\
  C3 & 3 & Dress \\
  C4 & 4 & Coat \\
  C5 & 5 & Sandal \\
  C6 & 6 & Shirt \\
  C7 & 7 & Sneaker \\
  C8 & 8 & Bag \\
  C9 & 9 & Ankle boot \\
  \hline
\end{tabular}
\caption{Corresponding class of MNIST and Fashion-MNIST}
\label{tb:digit_fashion}
\end{center}
\end{table}

\begin{figure}[!t]
	\centering
	\includegraphics[width=4in]{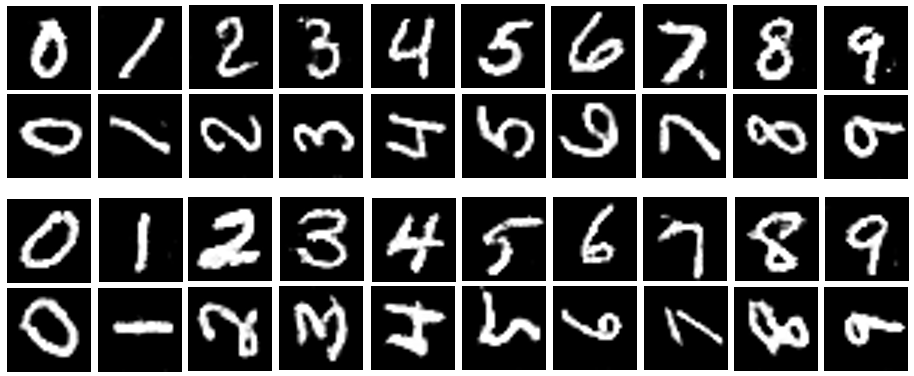}\\
	\caption{MNIST dataset with $0^\circ$ and $90^\circ$.}
    \label{sync_mnist_90_gen}
\end{figure}

\begin{figure}[!t]
    \centering
	\includegraphics[width=4in]{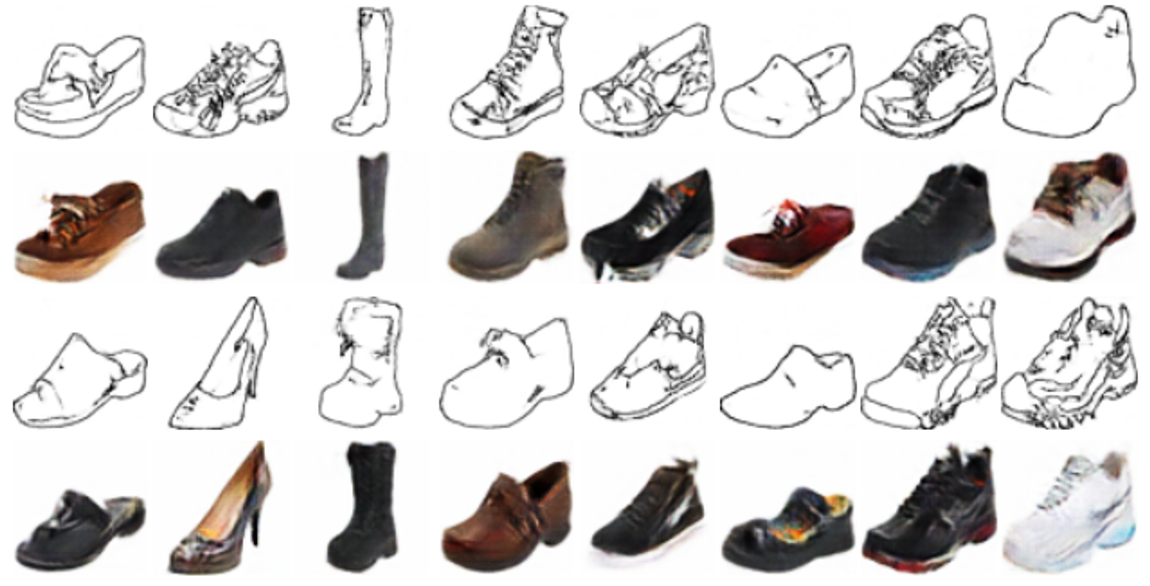}\\
	\caption{Synchronous generation of sketches and photos for shoes.}
    \label{shoes_gen}
\end{figure}

\subsection{Cross-domain Generation}
Our SyncGAN model can achieve good performance on cross-modal generation. In addition, it can also be applied to cross-domain generation for data with similar structures. 
We rotated 30000 images in the MNIST dataset by $90^\circ$ to create the image data in another domain. As shown in Fig.~\ref{sync_mnist_90_gen}, our model successfully generated synchronous images with $0^\circ$ and $90^\circ$ of images for the MNIST dataset. We also applied SyncGAN on the shoes images/sketches of the UT Zappos50K dataset~\cite{pix2pix2017}. 20000 pairs of cross-domain data were used to train the SyncGAN, and  Fig.~\ref{shoes_gen} shows the synchronous generation of shoes sketches/photos.

\subsection{Modality and Domain Transfer}
\begin{figure}[!t]
	\centering
	\includegraphics[width=4in]{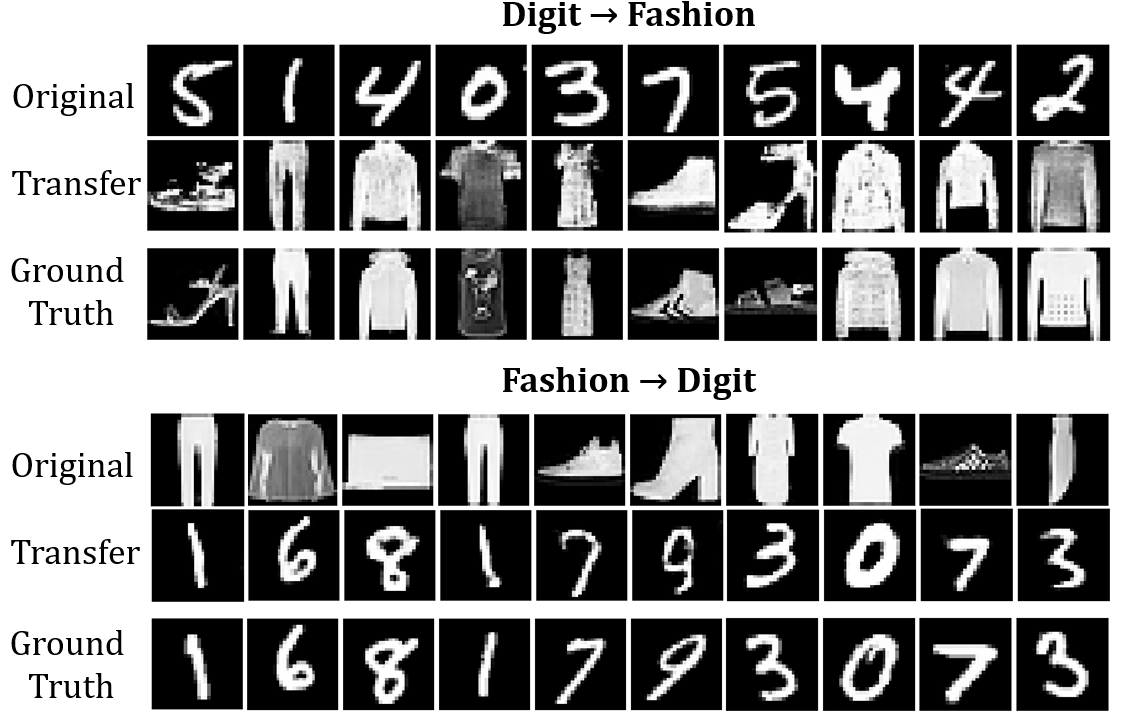}\\
	\caption{Bidirectional cross-modal transfer of MNIST and Fashion-MNIST.}
    \label{mnist_transfer}
\end{figure}

\begin{figure}[!t]
	\centering
	\includegraphics[width=4in]{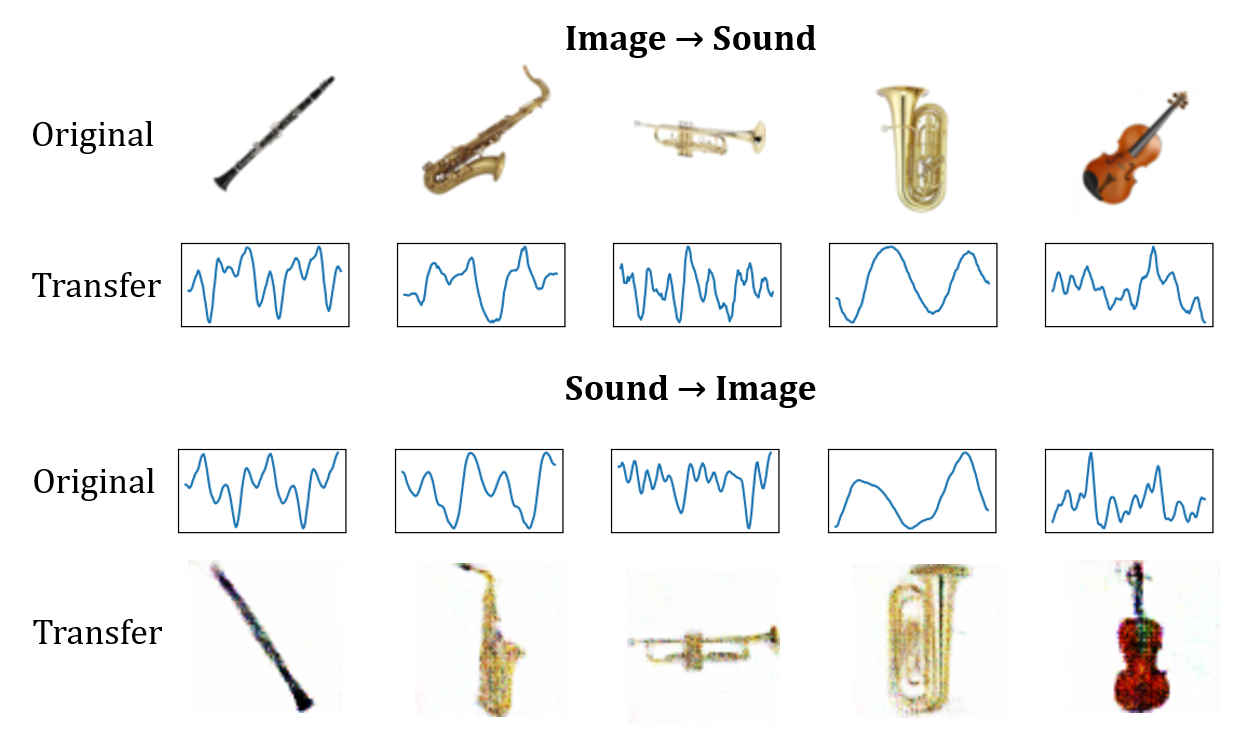}\\
	\caption{Bidirectional cross-modal transfer of images and sound for instruments.}
    \label{img_wav_transfer}
\end{figure}

\begin{figure}[!t]
	\centering
	\includegraphics[width=4in]{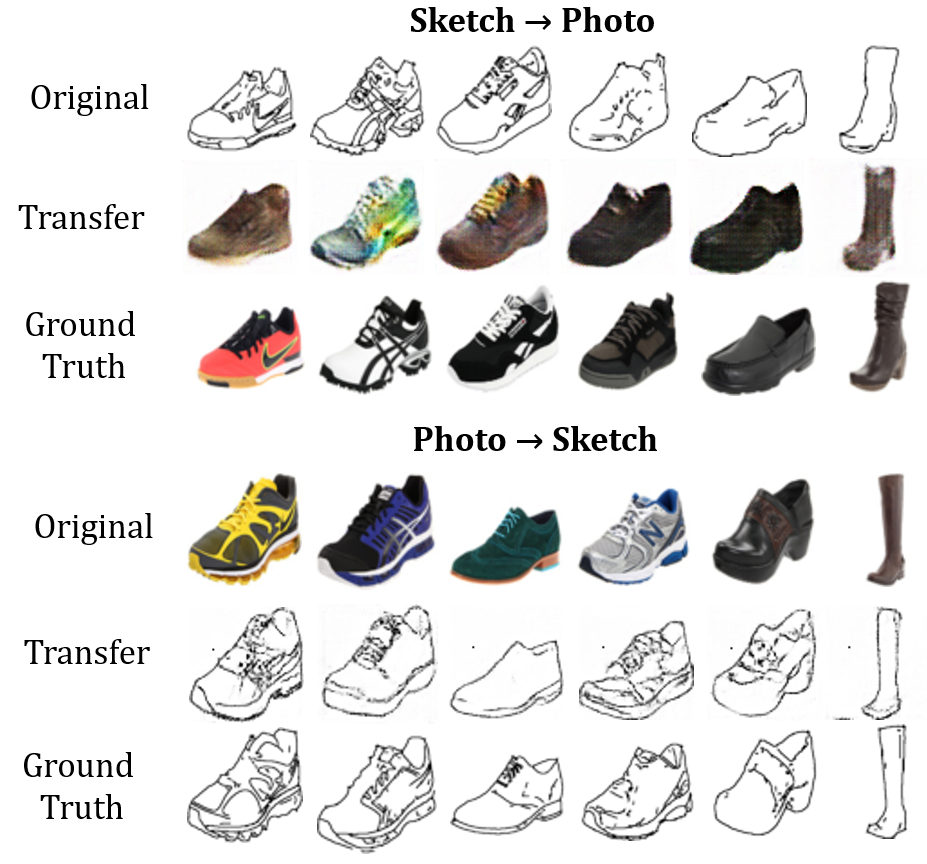}\\
	\caption{Bidirectional cross-modal transfer of sketches and photos for shoes.}
    \label{shoes_transfer}
\end{figure}

For the tasks of cross-modal and cross-domain transfer, we follow the method proposed by Lipton \textit{et al}.~\cite{lipton2017precise}, which reconstructs latent vectors by performing gradient descent over the components of the latent representations (Eq.~\ref{eq:inverse_map}).
\begin{eqnarray}
\label{eq:inverse_map}
z' \leftarrow z'-\eta \nabla_{z'}||G(z)-G(z')||_2^2
\end{eqnarray}
With the reconstructed latent vector and the generator of another modality/domain, we can successfully achieve bidirectional cross-modal or cross-domain transfer. Please refer to Fig.~\ref{mnist_transfer}~-~Fig.~\ref{shoes_transfer}.

\subsection{Evaluation of Synchronous Rate}
\begin{figure}[!t]
	\centering 
	\subfigure[]{
		\includegraphics[width=3.3in]{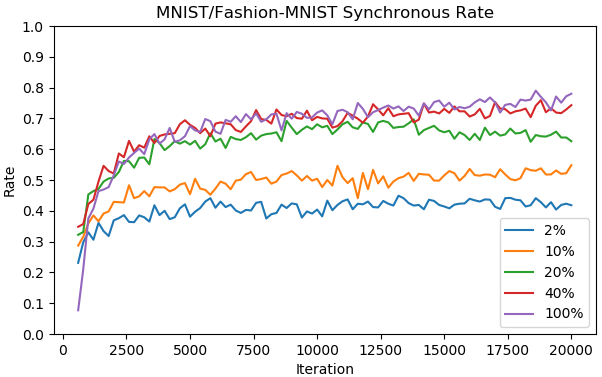}} 
	\subfigure[]{
		\includegraphics[width=3.4in]{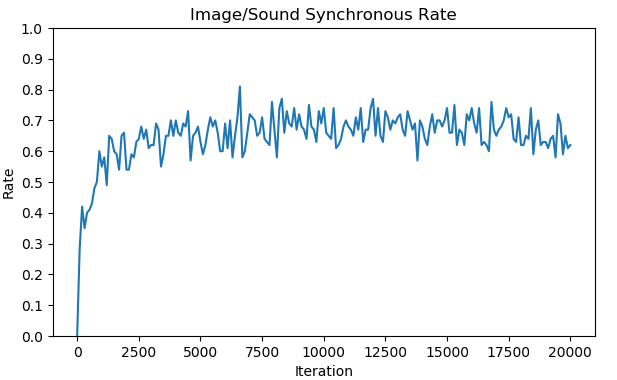}}
	\caption{Synchronous rates for (a) MNIST and Fashion-MNIST (b)images and sounds.}
	\label{sync_rate}
\end{figure}

For each cross-modal dataset, we trained a classifier for each modality based on data with concept labels. Given a cross-modal data pair generated by our SyncGAN, we examined whether the data pair is synchronous, i.e. two classifiers output the same labels. 
The synchronous rate is defined by the number of synchronous data pairs divided by the number of total generated data pairs. Fig.~\ref{sync_rate}(a) shows the synchronous rate for the MNIST dataset and the Fashion-MNIST dataset with different semi-supervised rate, which is defined by the number of paired data divided by the number of total data. The batch size in these experiments are 128.
We observed that using semi-supervised rate of 0.4 can achieve almost the same performance as supervised training (i.e. using semi-supervised rate of 1.0). Fig.~\ref{sync_rate}(b) demonstrates the synchronous rate of image and audio clip datasets. The result showed our model can be applied to real cross-modal data of image and sound, and the synchronous rate is about 0.7. The batch size in this experiment is 64.

\section{Conclusions and Future work}
\label{sec:conclusions}
Cross-domain GANs adopt several special mechanisms such as cycle-consistency and weight-sharing to extract the common structure of cross-domain data automatically. However, the common structure does not exist between most cross-modal data due to the heterogeneous gap. Therefore, the model need paired information to relate the different structures between data of various modalities which are of the same concept. In this paper, we present a novel network named synchronizer, which can constrain the latent space of generators in the GANs. Our SyncGAN model can successfully generate synchronous cross-modal/cross-domain data from identical random noises and perform transformation between different modalities/domains. Our model can generate synchronous image and audio data of instruments, and also can transfer data between these two modalities. However, the number of samples for each audio clip is only 512, which is around 0.01 secs. In the future, we will apply our model to sequential data such as longer audio clip and text.

\bibliographystyle{IEEEbib}
\bibliography{icme2018template}

\end{document}